\documentclass[utf8]{FrontiersinHarvard} 

\usepackage{url,hyperref,lineno,microtype}
\usepackage[onehalfspacing]{setspace}

\usepackage{booktabs}
\def\keyFont{\fontsize{8}{11}\helveticabold}
\def\firstAuthorLast{Guerra {et~al.}} 
\def\Authors{Alessio Guerra\,$^{1,2}$ and Oktay Karaku\c{s}\,$^{1,*}$}
\def\Address{$^{1}$Cardiff University, School of Computer Science and Informatics, Cardiff, UK.\\
$^{2}$Admiral Insurance, Ty Admiral, David St, Cardiff CF10 2EE, UK.
}
\def\corrAuthor{Oktay Karaku\c{s} - Cardiff University, School of Computer Science and Informatics, Abacws, Senghennydd Road, Cardiff, CF24 4AG, UK.}

\def\corrEmail{karakuso@cardiff.ac.uk}

\begin{document}
\onecolumn
\firstpage{1}

\title[Measuring Hope and Fear]{Sentiment Analysis for Measuring Hope and Fear from Reddit Posts During the 2022 Russo-Ukrainian Conflict} 

\author[\firstAuthorLast ]{\Authors} 
\address{} 
\correspondance{} 

\extraAuth{}

\maketitle

\begin{abstract}
This paper proposes a novel lexicon-based unsupervised sentimental analysis method to measure the ``\textit{hope}” and ``\textit{fear}" for the 2022 Ukrainian-Russian Conflict. \textit{Reddit.com} is utilised as the main source of human reactions to daily events during nearly the first three months of the conflict.  The top 50 ``hot" posts of six different subreddits about Ukraine and news (Ukraine, worldnews, Ukraina, UkrainianConflict, UkraineWarVideoReport, UkraineWarReports) and their relative comments are scraped and a data set is created. On this corpus, multiple analyses such as (1) public interest, (2) hope/fear score, (3) stock price interaction are employed. We promote using a dictionary approach, which scores the hopefulness of every submitted user post. The Latent Dirichlet Allocation (LDA) algorithm of topic modelling is also utilised to understand the main issues raised by users and what are the key talking points. Experimental analysis shows that the hope strongly decreases after the symbolic and strategic losses of Azovstal (Mariupol) and Severodonetsk. Spikes in hope/fear, both positives and negatives, are present after important battles, but also some non-military events, such as Eurovision and football games.

\tiny
 \keyFont{ \section{Keywords:} keyword, keyword, keyword, keyword, keyword, keyword, keyword, keyword} 
\end{abstract}

\section{Introduction}
For many years, the war in Europe has been just a dark memory. When on the 24$^{th}$ of February 2022, The Russian Federation declared war on Ukraine, it came out as a shock for most people all around the world \citep{faiola_2022}. It was thought that the presence of NATO and the European Union would be enough to guarantee peace in a short time. Unfortunately, that has been not the case due to the reason that both parties are neither part of NATO nor the EU, but are both former members of the USSR, and the conflict is still going on at early 2023.

In war, the morale of the nations is one of the most important elements \citep{pope1941importance}, since it is what pushes a country, most importantly a country that keeps fighting. In the case of a country defending its own land, the morale does not only regard the two-belligerent country but mostly the defenders. In fact, at first, the Ukrainian chance for success has been seen as tied to the support of western countries \citep{galston_2022}, the need that was also confirmed by the Ukrainian president himself \citep{france24_2022}. For this reason, the feelings of the western countries who support Ukraine could be a decisive factor in the future of the conflict. If the western audience would perceive the conflict as a lost battle, which, if dragged on, would have bad repercussions on their daily life and only cause more to Ukrainians, it could cause them to pressure their governments into stopping the support. On the other side, if there is the hope of winning the conflict, it is possible for the governments to keep guaranteeing active support to Ukraine and costly sanctions to Russia. 

According to the Collins dictionary, hope is an uncountable noun and is described as ``a feeling of desire and expectation that things will go well in the future" \citep{collins}. Conversely, fear is defined as ``a thought that something unpleasant might happen or might have happened" \citep{collins2}. As grammatical objects they may be uncountable nouns, however, the main purpose of this paper is to promote various text mining and sentimental analysis techniques to measure ``\textit{Hope}" and its negative counterpart ``\textit{Fear}" by using social media posts from Reddit.com - the social news aggregation, content rating, and discussion website. 

\section{Background \& Related Works}

From a general point of view, ``sentiment analysis" can be defined as the procedure of utilising important techniques such as natural language processing, text analysis and mining in order to extract and interpret subjective and human-related information. The source of information for sentiment analysis can be diverse e.g. written text or voice whilst the entities might be events, topics, individuals, and many more \citep{liu2020sentiment}. Sentiment analysis is also a broader name for many other tasks such as opinion mining, sentiment mining, emotion analysis and mining \citep{nasukawa2003sentiment, dave2003mining, liu2020sentiment}. Text data mining can be defined as the process of extracting data from structured and/or unstructured data mainly made of text \citep{hearst1999untangling}. Text mining can be utilised for different purposes and with many techniques like topic modelling \citep{rehurek2010software} and sentiment analysis \citep{feldman2013techniques}.  Text-related sentiment analysis is a versatile approach that helps to automatically extract meaningful information from the written text and useful to pursue many different objectives such as to assess and monitor psychological disorders \citep{zucco2017sentiment}, to evaluate human behaviours during the football World Cup 2014 \citep{yu2015world}, to detect emotions in general \citep{peng2021survey} or to use them to conclude on gender differences \citep{thelwall2010data}, or even to make predictions on the stock market \citep{pagolu2016sentiment} and measure heterogeneity of investors via their social media posts \citep{10.3389/frai.2022.884699}.

Considering vast amount of social networks recently continue to expand with regards to number of users, and are capable of reaching more audiences from nearly all levels of the community, Social media has naturally become the main source of information for text mining and sentimental analysis purposes. Sentimental analysis has been used to interpret data from different social network sources the most obvious example of which is Twitter \citep{hu2013unsupervised,yu2015world,giachanou2016like,10.3389/frai.2022.884699}. In addition, other popular social networks have also been used as the data source for the sentiment analysis realted purposes e.g. Facebook \citep{ortigosa2014sentiment}, Reddit \citep{melton2021public}, MySpace \cite{thelwall2010data} and even YouTube comments \citep{tripto2018detecting}.

Despite the social media being one of the most common sources of data, sentimental analysis has also found application basis for many more text corpora - to name but a few: movie \citep{thet2010aspect} or product reviews \citep{haque2018sentiment}, newspaper articles \citep{balahur2009rethinking}, or emails \citep{liu2018email}. Many of the analyses mentioned above mostly focus on understanding if a text is positive, negative, or neutral as a classifier \citep{pak2010twitter}, and/or promoting utilisation of various scoring systems \citep{naldi2019review}. It is also possible to employ similar analyses to understand if text utilises subjective or objective language \citep{liu2010sentiment}, or to interpret which emotions are conveyed \citep{yadollahi2017current}.

Having the vast amount of data containing multitude types of human emotions is not only highly exciting in terms of computational data analysis research, but also seen useful for the human behavioural research. In general, there are two main theories on how emotions are formed in the human brain. The first is the discrete emotion theory that says emotions arise from separate neural systems \citep{ekman2013emotion,shaver1987emotion}. In these seminal studies, \citep{ekman2013emotion} recognise 6 basic emotions of anger, disgust, fear, joy, sadness, and surprise whilst \citep{shaver1987emotion} recognise anger, fear, joy, love, sadness, and surprise. On the other hand, the dimensional model says that a common and interconnected neurophysiological system causes all effective states \citep{plutchik2013emotion,lovheim2012new}. In particular, \citep{plutchik2013emotion} recognise anger, anticipation, disgust, fear, joy, sadness, surprise, and trust whilst \citep{lovheim2012new} recognises anger, disgust, distress, fear, joy, interest, shame, and surprise. Creating statistical correlation and independence analysis approaches are also highly important to provide evidences for the aforementioned human behavioural studies. 

This paper aims to develop a novel lexicon-based unsupervised method to measure the ``hope” and ``fear" of the Ukrainian-Russian Conflict. Reddit.com is utilised as the main source of human reactions to daily events during nearly the first three months of the conflict. The structure of this social network - Reddit.com - allows for discussing about very specific topics (posting in specific subreddits), without short limitations on the number of characters that can be posted. This makes it easy to mine for opinions about the Ukrainian conflict, to get an idea for what people think about it and how hopeful/fearful they are. To achieve this goal, the top 50 ``hot" posts of six different subreddits about Ukraine and news (Ukraine, worldnews, Ukraina, UkrainianConflict, UkraineWarVideoReport, UkraineWarReports) and their relative comments are scraped and a data set is created. On this corpus, multiple analyses are employed. We promote using a dictionary approach, which scores the hopefulness of every submitted user post. The Latent Dirichlet Allocation (LDA) algorithm of topic modelling is also utilised to understand the main issues raised by users and what are the key talking points.

This research aims to fill the gap present in the literature regarding opinion mining, specifically for \textit{hope}. The main analysis consists of mapping hope measured with the newly proposed method. In particular, first, the trend of hope over the time is monitored. It is later compared with some of the most important events which happened during study time frame. This ascertains how such events influenced the public perception of the conflict, and provides evidence about the validity of the proposed hope measure. Fear is measured and mapped over the same study time period. In order to measure both fear and hope, a dictionary approach is employed that promotes using the National Research Council (NRC) Word-Emotion Association Lexicon dictionary as a starting point. Furthermore, individual topics extracted via the topic modelling observations are studied to interpret whether there is a correlation with ``hope" and what kind of relationship they present if this is the case. Sentiment analysis is also employed to track the popularity of individual leaders (Putin and Zelensky) and the Russian and Ukrainian governments. Finally, stocks such as Gazprom and indices (gas prices and Russian and Ukrainian bonds) are analysed to interpret whether there is a relationship between the developed hope score and the stock market.

%

\section{Methodology}
\subsection{Reddit Data}
Reddit has been chosen since its structure allows to easily group submissions about a specific topic, and because, compared to other social media platforms, the success of content is less influenced by the success of the author. To gather data for the analysis, it was necessary to obtain it from Reddit. The best way to achieve this goal is to use the official Reddit API. To do so it is necessary to register as a developer on their website, authenticate, register the app, state its purpose and functionality. Once the procedure is completed, the developer can request a token which has to be specified along with the client id, user agent, username, and password every time that new data is requested. 

Six subreddits were chosen for their relevance to the conflict:
\begin{itemize}
    \item r/Ukraine
    \item r/worldnews
    \item r/ukraina
    \item r/UkrainianConflict
    \item r/UkraineWarVideoReport
    \item r/UkraineWarReports
\end{itemize}	

The script developed in Python crawls the top 50 posts for each of the subreddit and the relative comments. Subsequently, it combines the new gathered submissions with the previously collected ones. It then removes eventual duplicates using the submission id. For every submission, the subsequent information was obtained:
\begin{itemize}
    \item title (only for posts): the title of the post
    \item text: the actual content of the submission
    \item upvotes
    \item author
    \item date
    \item id: the unique submission id
    \item flair: categorisation of the post by the author
    \item type: post or comment
    \item parent\_id
    \item subreddit
\end{itemize}

The data collection process started on the 10th of May and has been completed on the 28th of July. It was conducted daily around 3.00 pm UK time.
More than 1.2 million unique observations were gathered within this time frame.

\subsection{Pre-processing Stages}
The data obtained through the collection process was not useful on its own. It had to be processed to be analysed and explored. First, some cleaning needed to be done. Not all the observations gathered would be useful. In fact, some of the submissions in the r/worldnews subreddit were not about the conflict. To eliminate the irrelevant ones, only the posts with the flair ``Ukraine/Russia" had to be kept. The only issue is that flair is assigned only to ``post" type submissions, but not to comments.

Luckily, the structure of Reddit, allows to use id and parent\_id to move upwards to the original post from every comment. Every comment is like a tree branch in a forest-like structure, with every post representing a single tree. Thanks to this principle, it was possible to extract the ``ancestor\_id" of every submission and use it to assign a flair to the comments. This allowed to identify and remove the submissions without the relevant flair from the r/worldnews subreddit.

The next step would be converting all the words in each post to lowercase. Subsequently, we obtain score for a specific emotion for every submission. To reach this goal, the number of words related to the investigated emotion in every entry was counted. 

Another useful information to be extracted is the polarity score. Using a different sentiment analysis approach, the ``text" of a post or a comment would receive a score that ranges from -1 to 1 according to its sentiment. A score of -1 indicates a very negative meaning, while 1 indicates a very positive one. The score was extracted using the \textit{sentiment.polarity} method from the \textit{TextBlob} python module.
Another method, \textit{sentiment.subjectivity}, from the same module was also used that
allows us to understand if the author is stating facts or if they are voicing an opinion. Subjectivity ranges from a score of 0, which indicates a very subjective text, to 1, which indicates a very objective one.

One of the problems with dictionary-based sentiment analysis, is that it arbitrarily favours long texts. In fact, with a higher wordcount there are more chances to find the relevant words. Furthermore, it increases the score cap for a submission. A one-word comment could have a maximum score of one, while a hundred-words comment could potentially score one hundred. To solve this issue, a new parameter called ``$w_{lenght}$" was created. It stores the emotion score divided by the length of the submission multiplied by 100.
\begin{align}\label{equ:length}
    w_{lenght} = \dfrac{N_{emotion}}{length\times100}
\end{align}

Another improvement to be made regarded the weight of singular opinions. There are opinions which are more popular than others. On Reddit, it is easy to understand whether one post is popular by looking at the number of upvotes. To have a better understanding of the public opinion, it was relevant to weight the hope score to the number of upvotes. While being an improvement, simply multiply the ``$w_{lenght}$" score for the number of upvotes, would disfavour popular comments in unpopular posts. A very successful post would have a very high number of visualisations, comments and upvotes. A comment X, viewed by 100 people and upvoted by 10 (10\%) would have a higher score than a comment Y, viewed by 10 people and upvoted by 5 (50\% of viewers). To solve this issue, the number of upvotes needed to be weighted on the number of comments on a post, to obtain its relative popularity (opposed to the absolute one). A parameter storing the number of comments for every post (``$sub_{in-post}$") was obtained by counting the submissions for every ``ancestor\_id". Finally, another parameter ``$w_{upvotes}$" was created. It stores the value that ``$w_{lenght}$" multiplied by the number of upvotes divided by ``$sub_{in-post}$".
\begin{align}
    w_{upvotes} = \dfrac{w_{length}\times upvotes}{sub_{in-post}}
\end{align}

\begin{figure}[htbp]
    \centering
    \includegraphics[width=\linewidth]{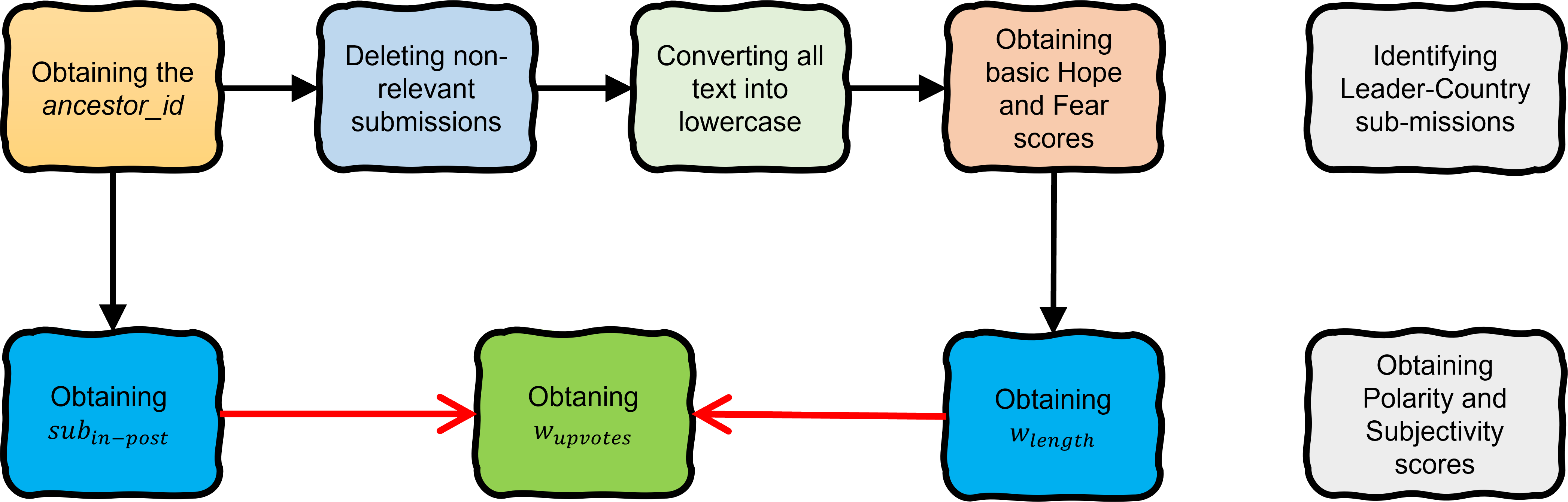}
    \caption{The pre-processing workflow}
    \label{fig:preprocess}
\end{figure}

Hence, $w_{upvotes}$ becomes the emotion score that is weighted on its length, the upvotes and the relative popularity. The flow diagram of the general pre-processing process is depicted in Figure \ref{fig:preprocess}.

\subsection{Measuring Hope and Fear}
Overall interest in the conflict has been measured in two different ways: (i) the number of submissions and (ii) the popularity of the posts. For the former, data were grouped by each day, and the number of daily submissions was counted. This includes both posts and comments, giving a good idea of the engagement trend. The latter studies the daily average number of upvotes for each post. Comments were excluded since a popular post is likely to host many comments with just one upvote, which would significantly lower the average. To achieve this goal, a post-only database was created. Data were grouped by date and the mean value for upvotes was computed.

Complementing the aforementioned second method with the first one is very useful to give a proper idea of the general interest trend. The number of posts could have been influenced by a small number of users who are somewhat involved with the conflict, while the public might not be this interested. This can be tested by looking at the popularity of the posts. In fact, popular posts have many upvotes. To reach them, submission needs to have the likeness or the attention of a big group of users.

The main goal of this dissertation is to map hope in western public opinion for the Russo-Ukrainian war. There is a gap in the literature regarding this specific issue. There is, indeed, no scholarly accepted way to automatically measure hope.

There are many ways to tackle sentiment analysis, like machine learning or dictionary-based approaches. The first one would have required labelling a dataset, saying what is hopeful and what is not. To properly do that, linguistic expertise is a requirement. On the other side, using a dictionary-based approach would allow using scholarly accepted dictionaries. Hence, this paper concerns a dictionary-based approach. 

Two issues had to be addressed to complete a dictionary-based analysis: that are linguistic and technical ones. At this point, we ask several important questions: What is hope and how do we measure it? According to the Collins dictionary, ``Hope is a feeling of desire and expectation that things will go well in the future”. Picking apart this definition helps to understand what are the elements that construct hope. The keywords are ``feeling”, ``well” and ``expectation in the future”. A feeling is something inherently subjective to the person who feels them. Well, in this case, indicates a positive outcome. The expectation is ``something looked forward to, whether feared or hoped for” and it is a synonym for anticipation.

Since there is no ``hope” dictionary to the best of our knowledge, one had to be developed. As a starting point, the NRC sentiment and emotion lexicon was used. The NRC Emotion Lexicon is a list of English words and their associations with eight basic emotions (anger, fear, anticipation, trust, surprise, sadness, joy, and disgust) and two sentiments (negative and positive). The annotations were manually done by crowdsourcing. Among the emotions catalogued in this dictionary, there is ``anticipation”, ``positive” and ``joy”. According to the previous definition, something to be hopeful needs to be subjective anticipation of a positive outcome. Hence, the three dictionaries were cross-referenced to find the words that showed ``anticipation” and at least one between ``positive” or ``joy”.

Thanks to this procedure, a ``hope” dictionary is developed. The lexicon respects two of the three parameters: ``anticipation” and ``positive outcome”. To satisfy the third one, all the Reddit submissions were analysed through the \textit{textblob.subjectivity} function. It gives a score that goes from 0 (not subjective) to 1 (very subjective). To respect the three parameters, only the submissions that present a minimum score of 0.5, are to be analysed.

Once the dictionary was developed, it needed to be implemented. Every submission is characterised by a ``text” column, which contains the message sent by the user. The script counts how many times words present in the ``hope” dictionary are also present in the ``text”. In this way, a raw hope score, notated as $hope_{score}$, is obtained, which is refined as described in the ``pre-processing” chapter of the paper

\begin{align}\label{equ:hope}
    hope_{score}=  \dfrac{\dfrac{N_{hope}}{lenght\times100}\times upvotes}{sub_{in-post}}
\end{align}

Fear was measured in the same way as hope. It is dictionary based and the score it is obtained by counting the fear-related words in every submission. The utilised dictionary was the same NRC one which is used to obtain “anticipation”, “joy” and “positive” words. The $fear_{score}$ is calculated as
\begin{align}\label{equ:fear}
    fear_{score}=  \dfrac{\dfrac{N_{fear}}{lenght\times100}\times upvotes}{sub_{in-post}}
\end{align}

\subsection{Leader and Country Analysis}
To obtain Leader analysis data, two new databases were created. The first one had only the submission containing the name “Zelenskyy” or its variations “Zelens'kyj” or “Zelensky”. The second one instead included only observations which presented the name “Putin”. Differently from the other analysis, hope and fear was not analysed, but the focus was on the sentiment polarity score. The polarity method from TextBlob was employed. It gives a score that ranges from -1 to 1, with the former representing a negative opinion, while the latter showing a positive one. After both databases were grouped by day, the mean daily polarity score was computed.

Similar to the Zelenskyy vs Putin analysis, two new databases were created. The first one included only submissions which contained the name “Ukraine”, while the second one only had only observations which presented the name “Russia”. Subsequently, the polarity score was measured using the TextBlob polarity method. Then, observations were grouped by day and the daily average polarity score was computed.

\subsection{Stock Market Analysis}
After collecting historical prices on six different stocks and financial titles (UK oil \& gas, Russian Ruble and US Dollar exchange rate, the price of gas and the price of crude oil), they were joined to the “daily” database. Said database contains the weighted average daily value for hope and fear.

We developed a linear regression model having the price of the ticker as the dependent variable and either the average weighted daily hope score or the weighted average daily fear score as the independent one. Then for each data set, we run this linear regression model and calculated the corresponding parameters for each modelling.

\subsection{Topic Modelling}
The aim of this analysis is to understand what the gathered submissions are about through topic modelling. Topic modelling is an unsupervised machine learning technique that allows us to organise, understand and summarise large bodies of text. It can be described as a method for extracting meaning out of the textual data by extracting groups of words, or abstract topics, from a collection of documents that best represents the information in the collection. More specifically, this technique returns a probabilistic distribution of different topics of discussion, where each topic is associated with a given document by a certain likelihood score. A document could be about different topics at the same time in different proportions.

We first created a corpus and dropped less frequent terms in it. Now that the text data have been processed, the optimal number of topics ($K$) is estimated. Using the \textit{searchK()} function, the different distributions of $K$ (from 2 to 10) are elaborated, so that it is possible to interpret the results and make a guess on the optimal number of topics into the model. To find the optimal number of topics, it is necessary to plot the distributions of K topics discovered according to various goodness of fit measures such as semantic coherence and exclusivity. Semantic coherence measures the frequency in which the most probable words in each topic occur together within the same document. Exclusivity on the other hand, checks the extent to which the top words for a topic are not top words in other topics. Coherence measure how a topic is strongly present and identifiable in documents, whilst exclusivity measures how much the topic differs from each other. The goal is to maximise both whilst keeping likelihood high and residuals low enough. Then the distribution of the topics in the document is examined to see if there is a prominence of one topic over the others or if they have similar distributions (bad sign). Subsequently, a word cloud for every topic is created. It shows in a graphical cloud all the top words, with size changing according to the relative frequency of the words. Using the \textit{labelTopics()} function, the words that are classified into the topics to better read and interpret them are inspected. This function generates a group of words which summarise each topic and measures the associations between keywords and topics. The most representative documents for each topic are then extracted. This is useful because it helps to give a more concrete idea of what each topic is about, using a real review as an example. The relationship between metadata, and topics is investigated. It is done defining the correlation model applying the \textit{estimateEffect()} function. This function performs a regression that returns the topic proportions as outcome variable. The output of the function has the aim to demonstrate the effect of the covariates of the topics. To conclude, the correlation between topics is studied.

\section{Experimental Analysis}
\subsection{Hope-Fear Analysis}
\begin{figure}[htbp]
    \centering
    \includegraphics[width=0.75\linewidth]{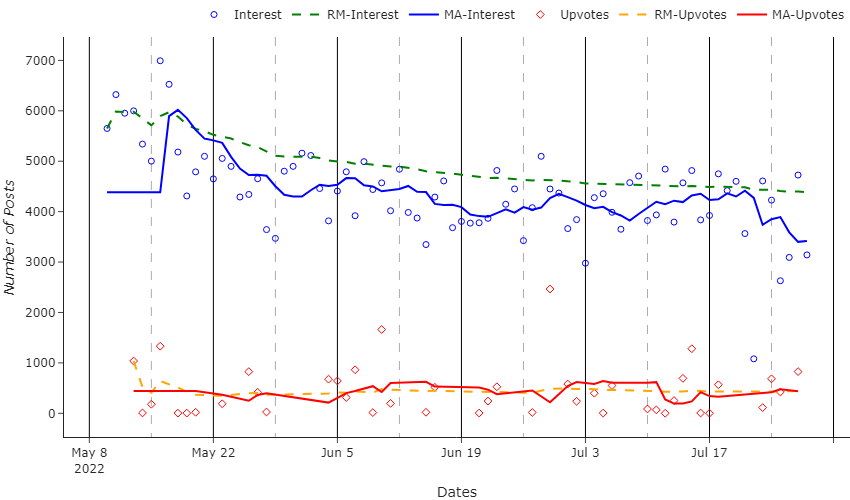}
    \caption{Number of submissions \& daily average number of upvotes over the time}
    \label{fig:interest}
\end{figure}

Our Hope-Fear analysis starts by measuring the public interest about the war and their intention to share posts in social media as shown in Figure \ref{fig:interest}. Overall social media interest during the conflict has been slowly but steadily decreasing for the whole analysed time window. With an average of 4335 daily submissions, in the first days, there were plenty of submissions, with a peak of 6993 posts in one single day on the 16th of May 2022. In the last part of the explored time, numbers are lower, with a negative peak of only 1080 submission in one day on the 22nd of July 2022, 5919 less than its maximum.

When we evaluate the daily upvote rates in Figure \ref{fig:interest}, differently from the above analysis, there are no significant changes in the trend of number of upvotes over time. The daily average itself is very volatile, but the trend remains stable.
This could mean that while the users are still receptive and supportive towards the Ukrainian conflict (they keep upvoting the most important posts), they are less engaged, posting and commenting less.

\begin{figure}[htbp]
    \centering
    \includegraphics[width=0.75\linewidth]{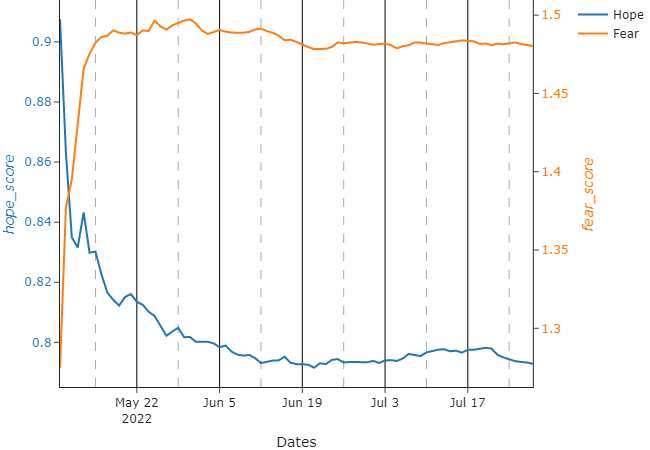}
    \caption{Running means for the proposed hope and fear scores}
    \label{fig:hope-fear}
\end{figure}

Thanks to this steady trend in upvotes and number of posts in each day, we calculated daily hope score by using the expression given in (\ref{equ:hope}). 
As it is possible to observe from the graph given in Figure \ref{fig:hope-fear}, the hope score during the analysed time-period is decreasing and finds a nearly-steady state after the half of the observed period in terms of its running mean visualisation. After the initial big drop, the score seems to stabilise on a lower value. This seems to reflect what happens during the war. In fact, the big drop happens around the fall of Azovstal (Mariupol) and Severodonetsk. Successively, it mirrors the “phase two” of the Russian offensive, with very slow and steady trend of hope score. This is also reflected by the fact that central 50\% of the observations of the hope score are in a range of 0.054, while the total range is 0.264, as it is possible to see from the descriptive statistics in Table \ref{tab:stats}.

\begin{table}[ht!]
    \centering
    \begin{tabular}{p{2.5cm}p{1.5cm}p{1.5cm}p{1.2cm}p{1.2cm}p{1.75cm}p{1.5cm}p{1.5cm}p{1.5cm}}
    \toprule
       	&\textbf{Interest}	&\textbf{Upvotes}	&\textbf{Hope}	&\textbf{Fear}	&\textbf{Zelenskyy}	&\textbf{Putin}	&\textbf{Ukraine} 	&\textbf{Russia}\\\toprule
Days	&80	&43	&81	&81	&81	&81	&81	&81\\
Mean	&4383.06	&441.38	&0.7928	&1.4803	&0.0949	&0.0381	&0.0901	&0.0402\\
St. Deviation	&861.06	&518.90	&0.0425	&0.0591	&0.0417	&0.0162	&0.0129	&0.0108\\
Minimum	&1080.00	&1.00	&0.6434	&1.2749	&-0.0311	&-0.0053	&0.0627	&0.0067\\
25$^{th}$ percentile	&3840.50	&20.00	&0.7650	&1.4405	&0.0716	&0.0254	&0.0813	&0.0359\\
50$^{th}$ percentile	&4412.50	&253.00	&0.7926	&1.4848	&0.0959	&0.0390	&0.0896	&0.0417\\
75$^{th}$ percentile	&4815.25	&659.50	&0.8189	&1.5213	&0.1206	&0.0476	&0.0991	&0.0480\\
Maximum	&6993.00	&2464.67	&0.9075	&1.6199	&0.2138	&0.0728	&0.1210	&0.0596\\\bottomrule
    \end{tabular}
    \caption{Descriptive statitics for the whole analysis}
    \label{tab:stats}
\end{table}

Similarly, by using the expression developed in (\ref{equ:fear}), we calculated the fear score for the same time period. Despite being pretty volatile, fear remains stable for the whole analysis just after inital couple of days. This is an interesting observation, especially when compared to hope, which decreases in the same time period. Hope-Fear results are slightly negatively correlated, with a Pearson correlation index of -0.986. Here, in order to clearly interpret this phenomenon, we plot running means of Hope and Fear on the same axes below in Figure \ref{fig:hope-fear}.

\subsection{Validation of Hope/Fear Scores}
In order to validate and better visualise the proposed hope/fear scores, we investigated 18 important events within the experimental period. To reach this, observations were grouped by day and the mean hope score was computed. The overall mean of the hope score was also calculated and a new column which contained the overall mean - each day average was created. The said important events chosen for the validation analysis are given below:
\begin{enumerate}
    \item \textbf{May 9} - failed Russian Donetsk River crossing. Ukrainian sources declare that during the crosses, 70 heavy Russian units were destroyed or lost.
    \item \textbf{May 13} - American-Russian talks. Lloyd Austin (American secretary of defence) and Sergei Shoigu (Russian minister of defence) held telephone talks for the first time since the start of the invasion.
    \item \textbf{May 15} - Ukraine won the Eurovision song contest thanks to an overwhelming popular vote. Stefania by the Kalash Orchestra won with 192 votes from the jury (4th place) and 439 from the televote. Second place went to the United Kingdom with 466 total votes.
    \item \textbf{May 17} - Azovstal, the steel factory of Mariupol is lost. It was the last stand of the Azov Battalion, a controversial group, which contained many of the best trained Ukrainian soldiers. This deprived Ukraine of a strategically important port, many soldiers and allowed the Russians to unify the front.
    \item \textbf{May 27} - 90\% of Severodonetsk is destroyed. The city is of big strategic importance, since it could allow the Russian to encircle many Ukrainian units in Donbass.
    \item \textbf{May 29} - First visit of Zelenskyy outside of Kiev. This visit had the purpose to show that the president was not afraid of Russia taking him out.
    \item \textbf{May 30} - Russian troops enter Severodonetsk
    \item \textbf{June 5} - Ukraine is eliminated in the World Cup qualifiers, after losing 1-0 to Wales, with a goal scored by Gareth Bale.
    \item \textbf{June 12} - Ukrainian supplies and planes destroyed.
    \item \textbf{June 16} - sinking of a Russian ship. The Pastel Vasily Bekh tug was sunk near snake island in the black sea.
    \item \textbf{June 17} - Putin speech at economic forum in St. Petersburg.
    \item \textbf{June 22} - Ukrainian drone strike on a Russian oil refinery.
    \item \textbf{June 26} - 14 missiles hit Kiev, damaging several buildings and a kindergarten.
    \item \textbf{July 6} - Russian duma prepares to go into war economy, which would allow to order companies to produce war supplies and make workers work overtime.
    \item \textbf{July 7} - Zelenskyy gave a speech on the effectiveness of western artillery. Furthermore, a technical pause from the Russian offensive started, with the aim to regroup.
    \item \textbf{July 14} - start of the volunteer mobilisation, which requires by the end of the month, 85 federal areas to recruit 400 men each.
    \item \textbf{July 16} - US house of representative approves a bipartisan bill that would grant \$100 million in funds to train Ukrainian pilots to fly US fighter jet.
    \item \textbf{July 23} - 4 Kalibr missiles hit Odessa. Of those 4, 2 were intercepted. The other 2 according to Russian sources destroyed a warship and a warehouse containing missiles.
\end{enumerate}

\begin{figure}[htbp]
    \centering
    \includegraphics[width=\linewidth]{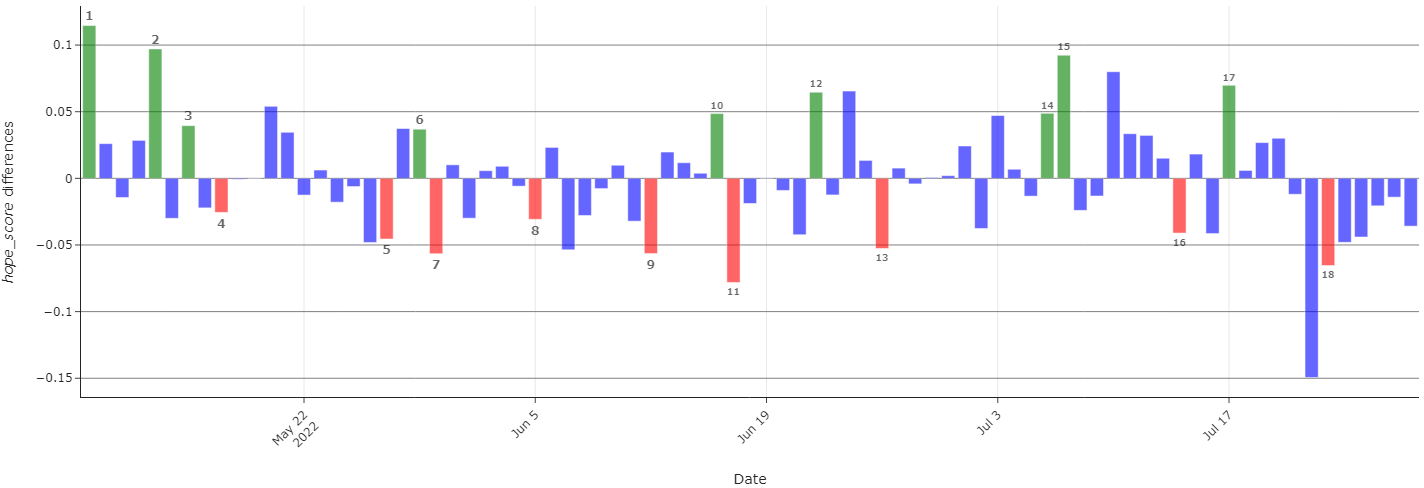}
    \caption{Deviation from the average hope.}
    \label{fig:mainevents}
\end{figure}

The graph in Figure \ref{fig:mainevents} shows how much above, or below average hope scored during the analysed period. Many of the spikes, both negative and positive, coincide with real world events which had an impact on the war or on the morale of the western public opinion. Some of the positive events include but are not limited to: the Ukrainian victory in the Eurovision contest (3), financial help packages from the United States (17) and the sinking of Russian ships (10). Negative ones include but are not limited to the loss of Azovstal (4), the fall of Severodonetsk (5) and the elimination of Ukraine from the World Cup 2022 (8). 

As it is possible to observe in Figure \ref{fig:mainevents}, most of the biggest positive spikes are concentrated in the first days, when the phase 2 of the war had recently started. After the fall of Azovstal and Severodonetks, a slower and more intense phase of the war starts. Russians advance slowly but steadily. This is also reflected in the graph, where we can observe few spikes, and many observations being below average for the whole duration of June. In July there was more movement, in fact the United States developed a plan of military and financial aid to Ukraine. Furthermore, Turkey managed to broker a trade deal between Ukraine and Russia, which would allow Ukraine to export grain, avoiding famine in many countries (mainly in Africa). At the same time, Russian advance keeps preceding recklessly, as shown by the negative spikes at the end of the month.

\subsection{Country-Leader Analysis}
\begin{figure}[htbp]
    \centering
    \includegraphics[width=0.49\linewidth]{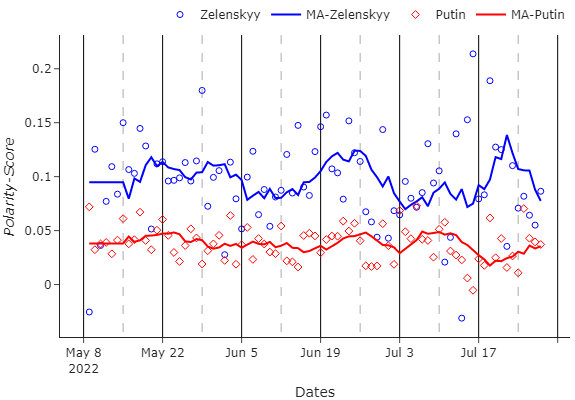}
    \includegraphics[width=0.49\linewidth]{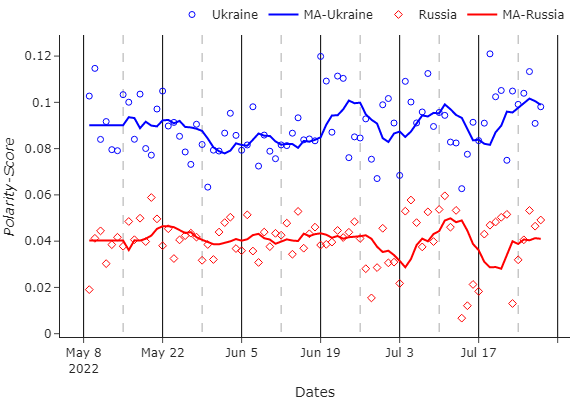}
    \caption{(LEFT) Polarity score for the two leaders. (RIGHT) Polarity score for the two countries. MA graphs for each figure refer to 7-days moving average of the original data.}
    \label{fig:country-leader}
\end{figure}

In this case of the experiments, we try to measure public interest in countries (Ukraine - Russia) and leaders (Zelenskyy - Putin). As previously stated, the metric for popularity refers to the sentiment ``polarity". The first and most obvious consideration that emerges from this analysis presented in Figure \ref{fig:country-leader} is that Zelenskyy, the president of Ukraine, presents a higher sentiment than Putin, president of Russia. As it is possible to notice, Zelenskyy is consistently more popular than his Russian counterpart, for the whole analysed period. In fact, the average polarity score for the Ukrainian president is 0.097, 2.6 times more than Putin, who scores a mere 0.037.
Despite being less popular, the Russian president is more interesting to the Reddit community than Zelenskyy. In fact, his name is cited 30663 times in the database, 7.2 times more than his Ukrainian counterpart, who is cited only 4055 times.

Another interesting point is that despite being relatively volatile, the trend seems to be consistent during the analysed period. None of the two leaders present an increase, nor a decrease, in popularity. Zelenskyy shows a higher volatility than Putin, but this is likely attributable to the smaller sample size.

The small sample size also causes the big outliers in the Zelenskyy graph. For example, on the fourteenth of July 2022, the Ukrainian president shows a polarity score of -0.31, 0.128 below the average score. There are only 49 submissions naming Zelenskyy on that day. One of the first ones, accuses the president to be a Nazi and to have violated human rights in Donbass. Many comments answer to these accusations defending the president. Saying for example:
\begin{center}
``\textit{this is such a massive false equivalence. periodically i bother responding to it.  here is my copy-paste nobody ever wants to engage with. non-extensive list examples of ways in which i think it’s possible to differentiate the two cases:* zelensky has never used chemical weapons to suppress a revolt against his rule by an ethnic minority, * the us did not execute civilians en mass in any captured town [...]}"
\end{center}
or:
\begin{center}
``\textit{this is a ludicrous comparison. whilst i don’t agree with what the west did in iraq in early 2000’s ….sadam hussein was committing genocide against the kurds, systematically slaughtering hundreds of thousands of people because of their race/religion. zelensky is not doing this, he is a democratically elected official and ukraine are a peaceful nation. so the idea that we (the west) are not allowed to comment on the russian invasion of ukraine because we’ve done something similar is lazy, ridiculous and without being rude to you, a tad stupid.}"
\end{center}

Most of those comments are saying that Zelenskyy and Ukraine did not commit atrocities, as affirmed by someone else. But (as it is later explained in the limitation part), many words with negative sentiment like ``suppress", ``execute", ``genocide", ``slaughtering", ``lazy", ``stupid" are used and the context is not interpreted. Having a big sample prevents these context-based exceptions from happening. For this specific day, the sample size is relatively small and is not able to counterbalance this single thread.

Another interesting insight is that there is basically no correlation between the popularity of Zelenskyy and Putin. The Pearson correlation index in fact is -0.03. It could have been possible to hypothesise a negative correlation between the two, maybe connected to the tides of the war. For example, if Russia was making gains Putin’s popularity could be increasing, while Zelenskyy’s would be decreasing. But this hypothesis is disproven by the evaluated data in the given time period.
This could be explained by the fact that it is possible that Putin’s popularity would not increase with a successful war, since he has mostly seen as the enemy.

Similar to the Putin vs Zelenskyy analysis above, it can be explored from the Figure \ref{fig:country-leader} that Ukraine scores evidently better than Russia. In fact, the former consistently more than the latter with an average polarity of 0.077, compared to an average of 0.044. In the same fashion of the previous analysis, Russia is cited way more frequently than Ukraine. In fact, the former is cited 137419 times, whilst the letter is 89736. This is found pretty interesting since, despite five of the six analysed subreddits being named after Ukraine, the real focus is Russia.

The two trends seem very similar. In fact, the Pearson correlation is 0.55. This might be because the two countries are very often cited in the same submission, hence presenting identical polarity scores. To solve this issue, two new databases which respectively contained “Ukraine” but not “Russia” and vice versa are created. In this process, 33790 observations for each database were dropped, removing more than one third of the original ``Ukraine" database.

The new numbers highlight even more focus on Russia, who now counts almost double the number of citations than Ukraine, counting 103629 against 55946. The new data shows an increase in the gap between the two countries. In fact, Ukraine, with an average score of 0.09 scores more than double than Russia, which decreases its polarity to 0.04. As expected, also the Pearson correlation index decreases significantly to 0.26, which remains still surprisingly high.

\subsection{Stock Market Analysis}
Four different tickers, regarding four different aspects connected to the war, were chosen: (1) United Kingdom Oil and gas stock price, (2) Ruble – US Dollar exchange rate, (3) Oil price, and (4) Gas Price. In particular, the most influential one is gas prices which have been used as a leverage for a good chunk of the conflict. Many western countries, including but not limited to Italy and Germany, provide weapons and support to Ukraine, but used to rely heavily on Russian gas for their energy needs. Russia has manoeuvred the gas price and supply (for example closing the gas pipeline North Stream One) to try to weaken the support for the Ukrainians and lift the sanctions imposed on them. Furthermore, through the increase of gas price, Russia secured record earnings and export levels. As always, in the stock market, prices are not only a reflection of current demand and offer, but also the projected demand and offer in the future. For all those reasons, we found it interesting to explore if a relationship existed between the hope and fear towards the conflict and the price of gas.

Oil price was also chosen for similar reasons. Oil is another combustible fuel which can be used to produce electricity. If natural gas is to become scarce, it is one of the most likely substitutes for many usages. Furthermore, the quota controlled by Russia is not big enough to allow them to manipulate the prices in the same way they do with gas. Considering that the energy crisis could influence the perception of the conflict for the European public opinion, it is interesting to also explore the oil prices relationship with the proposed hope and fear scores.

One of the very first consequences of western sanctions on Russia, was the fall of the Ruble. Many speculations were done on how this would have affected the Russian economy and their ability to repay their debts. The matter became even more interesting when after it started to climb back, even reaching higher values than pre-conflict period. Since Russia sells a significant part of its gas in Rubles, the swinging of the value of the Ruble are very important to the Russian economy and they are not to be underestimated. The perception of the stability of the country, hence the trust of the market in its currency could be put in jeopardy by losing this war. This is a good reason to expand the study to the exchange rate between US dollar and Russian Ruble.

The United Kingdom has been one of the most supportive countries of Ukraine since the beginning of the war. Differently from Italy and Germany, they are not part of the European Union, and they have rich reserves of natural gas and oil. United Kingdom Oil and Gas is one of the main stocks for the British energy market. It could prove insightful to understand if there is a relationship between hope and fear towards the Ukrainian war and the stock price of a company which acts in a country involved in the war, is influenced by the price of gas and oil, but has access to national stocks and is less dependent on Russia.

We run a linear regression analysis between each of these stock market elements and the proposed hope/fear score. Evaluating the results, we conclude that in terms of $p$-value there was no significant correlation between hope/fear score and Oil-price, Ruble \& US dollar exchange rate, and UK Oil-Gas. 

\begin{figure}[ht!]
    \centering
    \includegraphics[width=0.52\linewidth]{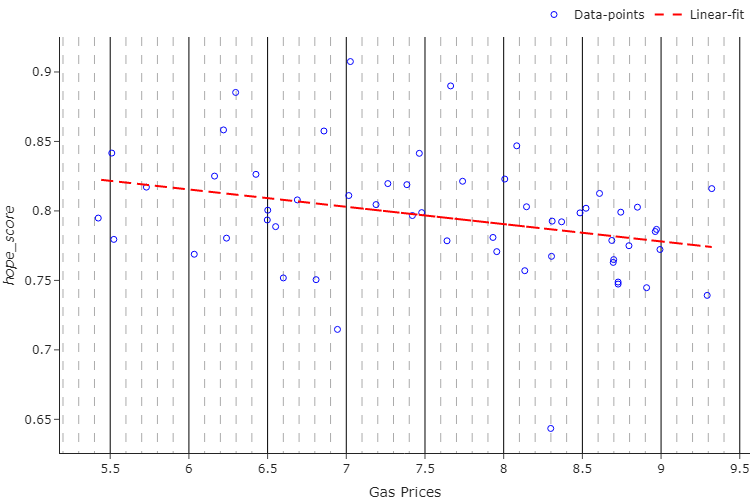}
    \includegraphics[width=0.47\linewidth]{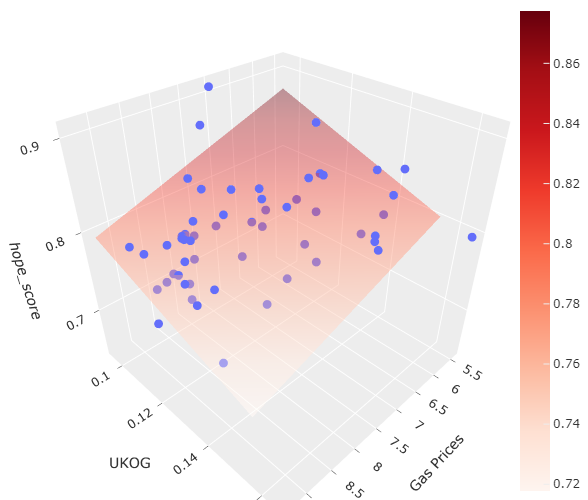}
    \caption{(Left) Scatterplot showing the gas price and the hope score. In red, the regression line. (Right) 3D-scatter plot of 2-regressor model fit.}
    \label{fig:stock}
\end{figure}

The similar insignificant relationship mentioned above was also obtained between fear score and gas prices. However, in terms of the hope score, a significant relationship was found between hope and the gas price. To interpret the relationship between the hope score and gas prices a linear regression was run, having the average daily hope score as the independent variable and the daily closing price as the dependent one. The regression presents a $p$-value of 0.018, showing the significance of the model whilst a relatively low $R^2$ value is obtained as 0.1. Furthermore, the Pearson correlation between the two variables is -0.32. As expected, the correlation is negative, so if hope goes up, the gas prices go down, or vice versa (See Figure \ref{fig:stock}-(Left)).

We also conducted a research on the relationship between the all stock variables as regressors and the hope/fear score as the target. Considering a significance threshold value of 0.05 for $p$-value, only the gas and UK Oil-Gas prices returned a significant relationship with the hope score whilst fear score does not provide a significant relationship with any of the regressors. Evaluating the results presented in Figure \ref{fig:stock}-(Right), we can conclude that a clear relationship between the hope score and two-regressor model (Gas\&OKOG) with $R^2$ value of 0.202 and again with a reciprocal proportion.

This analysis means that the public hope for the result of the conflict is not the primary driver for gas and UKOG prices, but there is indeed a relationship to be explored.

\subsection{Topic Modelling}
As is described in the previous sections, we now investigate the Reddit data set in terms of topic modelling. To achieve this goal, we utilised R programming language and many different R external packages are used:
\begin{itemize}
    \item \textbf{NLP:} provides the basic classes and methods for natural processing language and poses as a base for the following packages.
    \item \textbf{openNLP:} ``an interface to the Apache OpenNLP tools (version 1.5.3). The Apache OpenNLP library is a machine learning based toolkit for the processing of natural language text written in Java. It supports the most common NLP tasks, such as tokenization, sentence segmentation, part-of-speech tagging, named entity extraction, chunking, parsing, and coreference resolution \citep{apache}."
    \item \textbf{quanteda:} ``framework for quantitative text analysis in R. Provides functionality for corpus management, creating and manipulating tokens and ngrams, exploring keywords in context, forming and manipulating sparse matrices of documents by features and feature co-occurrences, analysing keywords, computing feature similarities and distances, applying content dictionaries, applying supervised and unsupervised machine learning, visually representing text and text analyses, and more \citep{quanteda}."
    \item \textbf{dplyr:} ``is a grammar of data manipulation, providing a consistent set of verbs that help to solve the most common data manipulation challenges \citep{dplyr}."
    \item \textbf{tidytext:} ``provides functions and supporting data sets to allow conversion of text to and from tidy formats, and to switch seamlessly between tidy tools and existing text mining packages \citep{tidytext}."
    \item \textbf{qdap:} ``automates many of the tasks associated with quantitative discourse analysis of transcripts containing discourse. The package provides parsing tools for preparing transcript data, coding tools and analysis tools for richer understanding of the data \cite{qdap}."
    \item \textbf{plotly} and \textbf{ggplot2:} are packages used for creating graphics for the analysis.
    \item \textbf{ggthemes:} is a package that enable better aesthetics for graphs.
    \item \textbf{wordcloud:} is a package that allows the creation of wordcloud-type graphs.
    \item \textbf{stm:} ``The Structural Topic Model (STM) allows researchers to estimate topic models with document-level covariates. The package also includes tools for model selection, visualisation, and estimation of topic-covariate regressions \cite{stm}". Structural Topic Modelling (STM) is a topic model method. It is a semi-automatic approach that allows us to incorporate metadata, which represents information about each document, into the topic model. STM aims at discovering topics, estimate their relationship to document metadata and gather information on how the topics are correlated.
\end{itemize}

\subsubsection{Estimating the optimal number of topics}

After the corpus is created, the first step is to extract the diagnostics and estimate the optimal number of topics. Whilst estimating the optimal number of topics, our aim is to maximise two important diagnostics of the \textit{exclusiveness} and \textit{coherence} whilst keeping \textit{likelihood} high and \textit{residual} diagnostics low enough. Due to the fact that having nine topics would ensure that there would be little mixing up between the topics, a little more importance is given to coherence. On the other hand, data would be very hard to interpret and would be difficult to extract useful information from it.

\begin{figure}[ht!]
    \centering
    \includegraphics[width=0.7\linewidth]{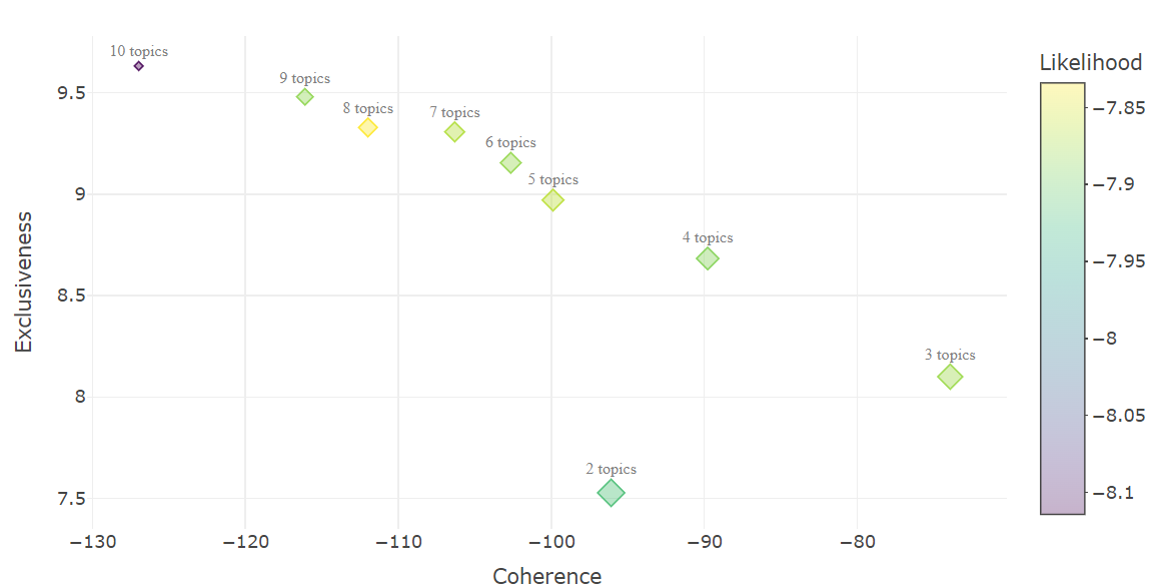}
    \includegraphics[width=0.28\linewidth]{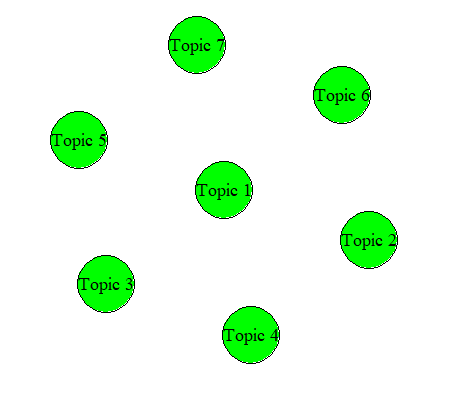}\\
    \includegraphics[width=0.44\linewidth]{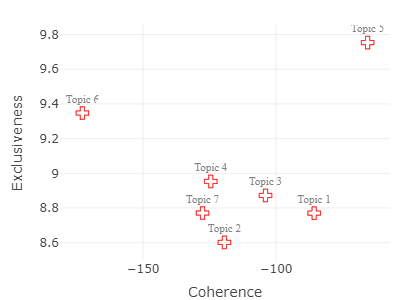}
    \includegraphics[width=0.55\linewidth]{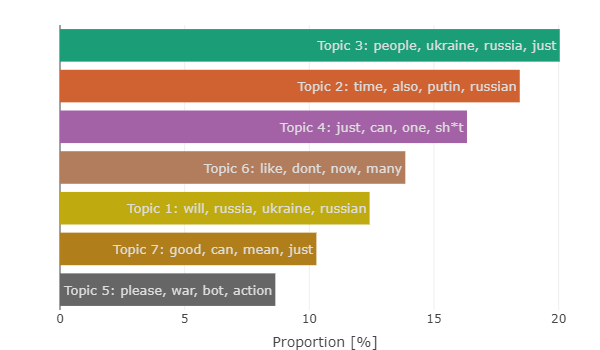}
    \caption{(a - Top Left) Model Selection results with four distinct diagnostics. Sizes of each marker relate to the residual diagnostic values. (b - Top Right) Exclusivity and coherence for the individual topics. (c - Bottom Left) Topic proportions in the dataset. (d - Bottom Right) Correlation between topics.}
    \label{fig:f15}
\end{figure}

We present the optimal number of topic selection diagnostic results in Figure \ref{fig:f15}-(a). Examining the Figure \ref{fig:f15}-(a), we can see that 7 and 8 number of topics appear to be the optimal choices as the result for the likelihood, residual and coherence-exclusiveness analysis. We stick with 7 number of topics as the optimal model since it has lower coherence value compared to 8 topics. Thus, two out of nine topics are discarded and seven is chosen as the topics for this analysis which are:
\begin{itemize}
    \item Topic 1: Geopolitical arguments
    \item Topic 2: Russia and government
    \item Topic 3: Morality of war
    \item Topic 4: War atrocities
    \item Topic 5: Submissions in Russian
    \item Topic 6: Foreign submissions
    \item Topic 7: Weapons
\end{itemize}

Examining Figure \ref{fig:f15}-(c), the quality of the topic is investigated in the same way as before, ideally coherence and exclusivity would be maximised. In this case it is possible to observe that Topic 5 greatly outperformed all the other topics, especially in coherence. This happens because those observations are all in Russian, this makes them very different from the rest. Topics 1 and 3 score very well on their own in terms of Coherence, whilst Topic 2 \& 7 are the worst performing ones overall. Topic 6 on the other side is the one that distinguishes itself the most in terms of exclusiveness, despite having a relatively low semantic coherence. The distribution of the topics is analysed in Figure \ref{fig:f15}-(d). Topic 3 is the most prominent topic, describing around 20\% of the database. Topics 5 and 7 are the less popular ones, scoring around 10\% each. Considering the correlation analysis plot in Figure \ref{fig:f15}-(b), we can clearly conclude that there appears to be no correlation between any of the topics.

\begin{table}[ht!]
    \centering
    \begin{tabular}{p{2.25cm}p{1.75cm}p{1.75cm}p{1.75cm}p{1.75cm}p{1.75cm}p{1.75cm}p{1.75cm}}
    \toprule
    \textbf{Results}
       	&\textbf{Topic 1}	&\textbf{Topic 2}	&\textbf{Topic 3}	&\textbf{Topic 4}	&\textbf{Topic 5}	&\textbf{Topic 6}	&\textbf{Topic 7}\\\toprule
Intercept	&0.1112	&0.1943	&0.2187	&0.1759	&0.0800	&0.1206	&0.0994
\\
$hope_{score}$	&0.0036	&-0.0035	&-0.0079	&0.0056	&-0.0040	&-0.0091	&0.0152
\\
$fear_{score}$	&0.0068	&-0.0051	&-0.0085	&-0.0098	&0.0060	&0.0137	&-0.0031
	\\\bottomrule
    \end{tabular}
    \caption{Topic Modelling Analysis Results}
    \label{tab:topic}
\end{table}

\subsubsection{Topic 1: Geopolitical arguments}
\begin{figure}[htbp]
    \centering
    \begin{tabular}{c|c|c}\textbf{Topic 1}&\textbf{Topic 2}&\textbf{Topic 3}\\
    \includegraphics[width=0.3\linewidth]{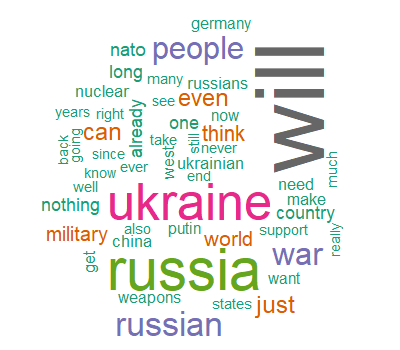}&
    \includegraphics[width=0.3\linewidth]{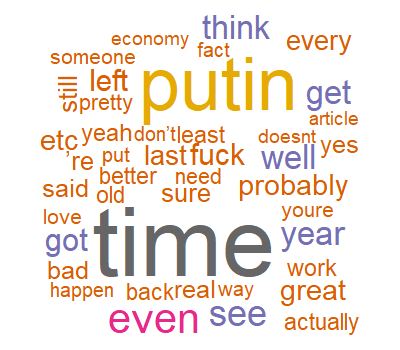}&
    \includegraphics[width=0.3\linewidth]{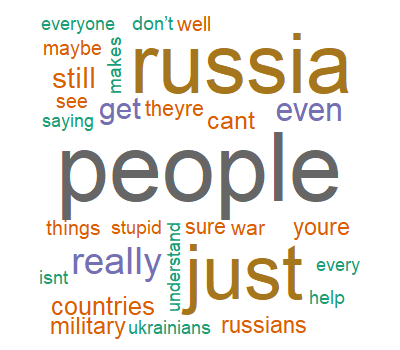}\\\bottomrule
    &&\\
    \textbf{Topic 4}&\textbf{Topic 5}&\textbf{Topic 6}\\
    \includegraphics[width=0.3\linewidth]{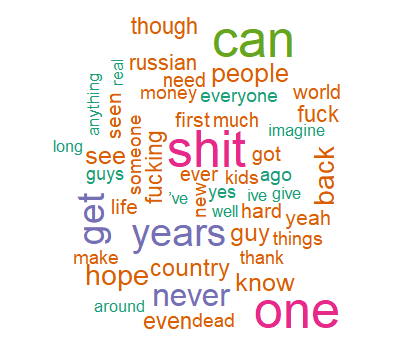}&
    \includegraphics[width=0.3\linewidth]{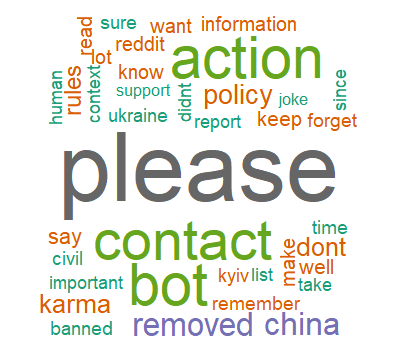}&
    \includegraphics[width=0.3\linewidth]{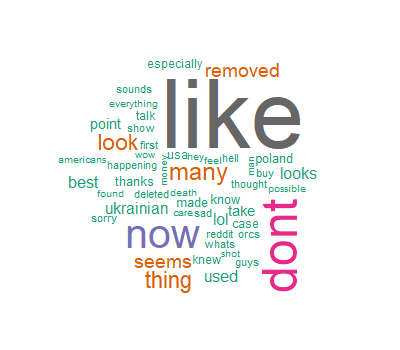}\\\bottomrule&&\\
    &\textbf{Topic 7}&\\
    &\includegraphics[width=0.3\linewidth]{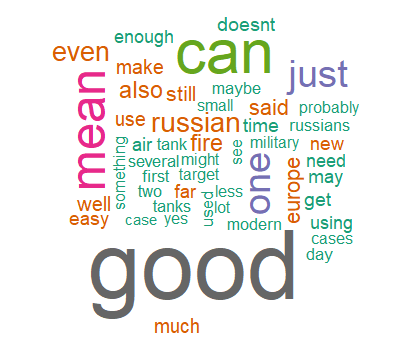}&\\
    \end{tabular}
    \caption{Wordcloud representation for each each topic.}
    \label{fig:cloud}
\end{figure}

In Table \ref{tab:topic}, linear regression modelling results of each topic with hope and fear scores are presented. It can bee seen that Topic 1 is positively correlated to both hope and fear. In addition, as shown  Fig \ref{fig:cloud}, Topic 1 is mostly about geopolitical argumentation. The most used words are ``Ukraine", ``Russia" and ``will", showing speculation about the conflict. Other popular words are ``NATO", ``china", ``Germany", ``support" and ``sanctions", a sign of how the broader picture is also depicted in the conversation. Furthermore, ``weapons", ``soldiers", ``nuclear" are also present, demonstrating an attention to battles.

The correlation to both hope and fear could be explained by the word ``will". If future possibilities are explored, they might be about positive events, hence increasing the hope score, or about scary ones, hence increasing the fear score.

\subsubsection{Topic 2: Russia and government}
Topic 2 is negatively correlated to both hope and fear. Topic 2 seems to be negative opinions about the Russians and governments. There are many words which refer to them as ``Putin", ``Russian", ``Russians", ``government", ``left" and ``right". Other popular words are ``f***", ``bad", ``wrong", ``f***ing", ``old" and ``stop". It is not very clear due to the low internal coherence of this topic.

\subsubsection{Topic 3: Morality of war}
Topic 3 is negatively correlated to both hope and fear. Topic 3 seems to be about the moral consequences of the war.  Investigating randomly taken submissions as examples shows us that the community discusses about (1) the morality of dealing economically with the side of the war, 
(2) the consequences positive of globalisation, and (3) the idea of leaving internal civic debates in Ukraine for later, while doing common front now against the common foe.

Being these moral considerations, they are not relevant with hope and fear, for this reason it is naturally considerable that they might score low in both.

\subsubsection{Topic 4: War atrocities}
Topic 4 is positively correlated with hope, but negatively with fear. Topic 4 is about war atrocities and their devastating effects. Unexpectedly, for this topic we obtained a positive correlation with hope and a negative one with fear. 

\subsubsection{Topic 5: Submissions in Russian}
Topic 5 is negatively correlated with hope, but positively with fear. Topic 5 is composed by the submissions in Russian language. It is negatively correlated to hope since there are no Russian words in the ``hope" dictionary. It is probably negatively correlated to fear because the few English words are present in the Fear dictionary (similar to the case in the third example).

\subsubsection{Topic 6: Foreign Submissions}
Topic 6 is negatively correlated to hope but positively correlated to fear. Similarly to the Topic 5, Topic 6 is mainly composed of submissions in foreign languages. Most of them will score 0 since their words will not be present in either dictionary. Potentially some similar common words in foreign languages with English created a positive correlation with Fear. 

\subsubsection{Topic 7: Weapons}
Topic 7 is positively correlated with hope, but negatively with fear. Topic 7 is about weapons. Many of the words shown reflect that: ``tanks", ``artillery", ``weapon", ``missiles", ``gun", ``range", ``modern", ``expensive", ``drone". Others also regard the military in a broader sense, like ``logistic", ``training" and ``equipment". Finally, ``good" is the most used word in the topic.

This explain that the superior Ukrainian equipment reassures the public and increases their hope.

\section{Discussions}
In Ukraine, many geopolitical themes are unfolding, and many interests are conflicting. Considering how high the stakes are, it is imperative for a good politician to use every tool at his disposal to direct the public opinion where it is most needed. 

Currently during the winter period in northern hemisphere, the stakes will be particularly high, since the electricity and gas demand will be particularly high, and likely gas price will be a strong weapon for Russia. Increased gas prices will have direct and indirect effects on prices for the public. Heating prices would go up significantly. The indirect effect would come from manufactured goods. In fact, with the increase in the electrical bill, would also come an increase of the price of their products. As for the analysed period, support towards Ukraine and Zelenskyy is still strong. The real test will be during the winter, when the western average Joe might be strongly affected by the consequences of the war, sometimes not even being able to afford heating and food. At this point it is possible that the public might ask to end the war at any cost. This might cause the end of European support in weapons and logistics, which would generate huge difficulties for Ukraine.

If the world of politics wants to keep support for the war, it might employ two strategies. The first one is to try to increase the hope of the people towards a Ukrainian victory. To do so, they should talk about the superiority of the weapons provided by the west, how good and effective they are compared to soviet era Russian ones. This topic in fact was positively correlated to hope and negatively correlated to fear.

The second one, would be to try to instate fear towards Russia. To do so, politics and news could start using those geopolitical arguments that prove the dangers of the country. This might be a double-edged strategy. In fact, it could undermine the faith in victory of the public and jeopardize overall morale. If this would happen and people would start to see Russia as an unstoppable danger, they could ask for a fast end of the conflict, since defeat would be seen as inevitable.

Another thing that would be deleterious are periods of excessive stagnation. They might cause a decrease in the interest, which coupled with possible severe economic consequences, might frustrate the public. The risk is that they would see his life worsened in exchange for no visible progress.

\section{Conclusions}
The results of this study can be seen as the development of a way to measure hope via exploiting social media posts of the public all over the world, and an insightful overview over the public opinion on the Russo-Ukrainian conflict, focused predominantly on hope.

The first analysis regards the interest towards the conflict. A steady decline in the number of submissions is observed, while the average number of upvotes for the posts does not increase or decrease. This shows a relative loss of interest, due to the stagnation of the news. In fact, the analysis takes place mostly during the “phase two” of the war, characterised by a slow but certain Russian advance. On the other side, the average number of upvotes remains constant, demonstrating that the potential interest is still present. The public is still there, it just needs something new to get engaged with and participate more actively again.

The second analysis is about hope. Following the events of the war, hope strongly decreases after the symbolic and strategical losses of Azovstal (Mariupol) and Severodonetsk. After that, it stabilises in its slow decrease, mirroring the tides of phase two of the conflict. Spikes in hope, both positives and negatives, are present after important battles, but also some non-military events, such as Eurovision and football games. This is an interesting insight, because it shows how morale is not only formed by the objective results of the war, but also by emotional events.

The third one regards fear. Its trend is stable during the entire analysis. Meaning that the tides of the war itself did not influence it significantly. There is a minor negative correlation with hope. It is interesting to notice that they are not inversely correlated. This means that hope and fear could coexist in the public opinion in specific instances.

The fourth one analyses the popularity of the two countries and their leaders, using a polarity score. The most obvious consideration is that Zelenskyy and Ukraine constantly outperform Putin and Russia. Despite being relatively volatile, the trend seems to remain constant. A key takeaway from this is that a strong opinion is formed, and without serious upheavals, it will not change.

In the fifth one, the relationship between fear/hope and relevant financial items is explored. A significant relationship (which is negative) between hope and the gas price was found. With the increase of hope, gas prices would decrease, or vice-versa. A reason for that could be that there is hope that a Ukrainian victory in the war would put ease again the gas flow from Russia to Europe. Since this has been selected as a fundamental analysis via limited amount of information, more studies would need to be done to fully explore this relationship.

The sixth one is the topic modelling. The submissions in English language are about five different topics: geopolitical arguments, Russia and government, morality of war, war atrocities and weapons. Those are the topics which caught the public eye the most in the analysed period. Geopolitical arguments are positively correlated with both hope and fear. Morality of war, Russia and government are negatively correlated with both hope and fear. Discussions about weapons are positively related to hope and negatively to fear, and surprisingly the same applies to war atrocities.

\bibliographystyle{Frontiers-Harvard}
\bibliography{test}

\begin{thebibliography}{41}
\providecommand{\natexlab}[1]{#1}
\expandafter\ifx\csname urlstyle\endcsname\relax
  \providecommand{\doi}[1]{doi:\discretionary{}{}{}#1}\else
  \providecommand{\doi}{doi:\discretionary{}{}{}\begingroup
  \urlstyle{rm}\Url}\fi
\providecommand{\selectlanguage}[1]{\relax}
\providecommand{\bibAnnoteFile}[1]{%
  \IfFileExists{#1}{\begin{quotation}\noindent\textsc{Key:} #1\\
  \textsc{Annotation:}\ \input{#1}\end{quotation}}{}}
\providecommand{\bibAnnote}[2]{%
  \begin{quotation}\noindent\textsc{Key:} #1\\
  \textsc{Annotation:}\ #2\end{quotation}}

\bibitem[{Balahur and Steinberger(2009)}]{balahur2009rethinking}
Balahur, A. and Steinberger, R. (2009).
\newblock Rethinking sentiment analysis in the news: from theory to practice
  and back.
\newblock \emph{Proceeding of WOMSA} 9, 1--12
\bibAnnoteFile{balahur2009rethinking}

\bibitem[{Benoit et~al.(2018)Benoit, Watanabe, Wang, Nulty, Obeng, Müller
  et~al.}]{quanteda}
Benoit, K., Watanabe, K., Wang, H., Nulty, P., Obeng, A., Müller, S., et~al.
  (2018).
\newblock quanteda: An r package for the quantitative analysis of textual data.
\newblock \emph{Journal of Open Source Software} 3, 774.
\newblock \doi{10.21105/joss.00774}
\bibAnnoteFile{quanteda}

\bibitem[{{Collins Dictionary}(2022{\natexlab{a}})}]{collins2}
{Collins Dictionary} (2022{\natexlab{a}}).
\newblock Fear definition and meaning: Collins english dictionary.
\newblock \emph{Fear definition and meaning | Collins English Dictionary}
\bibAnnoteFile{collins2}

\bibitem[{{Collins Dictionary}(2022{\natexlab{b}})}]{collins}
{Collins Dictionary} (2022{\natexlab{b}}).
\newblock Hope definition and meaning: Collins english dictionary.
\newblock \emph{Hope definition and meaning | Collins English Dictionary}
\bibAnnoteFile{collins}

\bibitem[{Dave et~al.(2003)Dave, Lawrence, and Pennock}]{dave2003mining}
Dave, K., Lawrence, S., and Pennock, D.~M. (2003).
\newblock Mining the peanut gallery: Opinion extraction and semantic
  classification of product reviews.
\newblock In \emph{Proceedings of the 12th international conference on World
  Wide Web}. 519--528
\bibAnnoteFile{dave2003mining}

\bibitem[{Ekman et~al.(2013)Ekman, Friesen, and Ellsworth}]{ekman2013emotion}
Ekman, P., Friesen, W.~V., and Ellsworth, P. (2013).
\newblock \emph{Emotion in the human face: Guidelines for research and an
  integration of findings}, vol.~11 (Elsevier)
\bibAnnoteFile{ekman2013emotion}

\bibitem[{Faiola(2022)}]{faiola_2022}
Faiola, A. (2022).
\newblock War in europe casts the continent into a frightening unknown.
\newblock \emph{The Washington Post}
\bibAnnoteFile{faiola_2022}

\bibitem[{Feldman(2013)}]{feldman2013techniques}
Feldman, R. (2013).
\newblock Techniques and applications for sentiment analysis.
\newblock \emph{Communications of the ACM} 56, 82--89
\bibAnnoteFile{feldman2013techniques}

\bibitem[{{France 24}(2022)}]{france24_2022}
{France 24} (2022).
\newblock Zelensky asks nato for weapons, west adds pressure on russia.
\newblock \emph{France 24}
\bibAnnoteFile{france24_2022}

\bibitem[{Galston(2022)}]{galston_2022}
Galston, W. (2022).
\newblock Opinion | give ukraine the weapons it needs.
\newblock \emph{The Wall Street Journal}
\bibAnnoteFile{galston_2022}

\bibitem[{Giachanou and Crestani(2016)}]{giachanou2016like}
Giachanou, A. and Crestani, F. (2016).
\newblock Like it or not: A survey of twitter sentiment analysis methods.
\newblock \emph{ACM Computing Surveys (CSUR)} 49, 1--41
\bibAnnoteFile{giachanou2016like}

\bibitem[{Haque et~al.(2018)Haque, Saber, and Shah}]{haque2018sentiment}
Haque, T.~U., Saber, N.~N., and Shah, F.~M. (2018).
\newblock Sentiment analysis on large scale amazon product reviews.
\newblock In \emph{2018 IEEE international conference on innovative research
  and development (ICIRD)} (IEEE), 1--6
\bibAnnoteFile{haque2018sentiment}

\bibitem[{Hearst(1999)}]{hearst1999untangling}
Hearst, M.~A. (1999).
\newblock Untangling text data mining.
\newblock In \emph{Proceedings of the 37th Annual meeting of the Association
  for Computational Linguistics}. 3--10
\bibAnnoteFile{hearst1999untangling}

\bibitem[{Hu et~al.(2013)Hu, Tang, Gao, and Liu}]{hu2013unsupervised}
Hu, X., Tang, J., Gao, H., and Liu, H. (2013).
\newblock Unsupervised sentiment analysis with emotional signals.
\newblock In \emph{Proceedings of the 22nd international conference on World
  Wide Web}. 607--618
\bibAnnoteFile{hu2013unsupervised}

\bibitem[{Ji and Han(2022)}]{10.3389/frai.2022.884699}
Ji, R. and Han, Q. (2022).
\newblock Understanding heterogeneity of investor sentiment on social media: A
  structural topic modeling approach.
\newblock \emph{Frontiers in Artificial Intelligence} 5.
\newblock \doi{10.3389/frai.2022.884699}
\bibAnnoteFile{10.3389/frai.2022.884699}

\bibitem[{Liu(2020)}]{liu2020sentiment}
Liu, B. (2020).
\newblock \emph{Sentiment analysis: Mining opinions, sentiments, and emotions}
  (Cambridge university press)
\bibAnnoteFile{liu2020sentiment}

\bibitem[{Liu et~al.(2010)}]{liu2010sentiment}
Liu, B. et~al. (2010).
\newblock Sentiment analysis and subjectivity.
\newblock \emph{Handbook of natural language processing} 2, 627--666
\bibAnnoteFile{liu2010sentiment}

\bibitem[{Liu and Lee(2018)}]{liu2018email}
Liu, S. and Lee, I. (2018).
\newblock Email sentiment analysis through k-means labeling and support vector
  machine classification.
\newblock \emph{Cybernetics and Systems} 49, 181--199
\bibAnnoteFile{liu2018email}

\bibitem[{L{\"o}vheim(2012)}]{lovheim2012new}
L{\"o}vheim, H. (2012).
\newblock A new three-dimensional model for emotions and monoamine
  neurotransmitters.
\newblock \emph{Medical hypotheses} 78, 341--348
\bibAnnoteFile{lovheim2012new}

\bibitem[{Melton et~al.(2021)Melton, Olusanya, Ammar, and
  Shaban-Nejad}]{melton2021public}
Melton, C.~A., Olusanya, O.~A., Ammar, N., and Shaban-Nejad, A. (2021).
\newblock Public sentiment analysis and topic modeling regarding covid-19
  vaccines on the reddit social media platform: A call to action for
  strengthening vaccine confidence.
\newblock \emph{Journal of Infection and Public Health} 14, 1505--1512
\bibAnnoteFile{melton2021public}

\bibitem[{Naldi(2019)}]{naldi2019review}
Naldi, M. (2019).
\newblock A review of sentiment computation methods with r packages.
\newblock \emph{arXiv preprint arXiv:1901.08319}
\bibAnnoteFile{naldi2019review}

\bibitem[{Nasukawa and Yi(2003)}]{nasukawa2003sentiment}
Nasukawa, T. and Yi, J. (2003).
\newblock Sentiment analysis: Capturing favorability using natural language
  processing.
\newblock In \emph{Proceedings of the 2nd international conference on Knowledge
  capture}. 70--77
\bibAnnoteFile{nasukawa2003sentiment}

\bibitem[{Ortigosa et~al.(2014)Ortigosa, Mart{\'\i}n, and
  Carro}]{ortigosa2014sentiment}
Ortigosa, A., Mart{\'\i}n, J.~M., and Carro, R.~M. (2014).
\newblock Sentiment analysis in facebook and its application to e-learning.
\newblock \emph{Computers in human behavior} 31, 527--541
\bibAnnoteFile{ortigosa2014sentiment}

\bibitem[{Pagolu et~al.(2016)Pagolu, Reddy, Panda, and
  Majhi}]{pagolu2016sentiment}
Pagolu, V.~S., Reddy, K.~N., Panda, G., and Majhi, B. (2016).
\newblock Sentiment analysis of twitter data for predicting stock market
  movements.
\newblock In \emph{2016 international conference on signal processing,
  communication, power and embedded system (SCOPES)} (IEEE), 1345--1350
\bibAnnoteFile{pagolu2016sentiment}

\bibitem[{Pak and Paroubek(2010)}]{pak2010twitter}
Pak, A. and Paroubek, P. (2010).
\newblock Twitter as a corpus for sentiment analysis and opinion mining.
\newblock In \emph{Proceedings of the Seventh International Conference on
  Language Resources and Evaluation (LREC'10)}
\bibAnnoteFile{pak2010twitter}

\bibitem[{Peng et~al.(2021)Peng, Cao, Zhou, Ouyang, Yang, Li
  et~al.}]{peng2021survey}
Peng, S., Cao, L., Zhou, Y., Ouyang, Z., Yang, A., Li, X., et~al. (2021).
\newblock A survey on deep learning for textual emotion analysis in social
  networks.
\newblock \emph{Digital Communications and Networks}
\bibAnnoteFile{peng2021survey}

\bibitem[{Plutchik and Kellerman(2013)}]{plutchik2013emotion}
Plutchik, R. and Kellerman, H. (2013).
\newblock \emph{Emotion, psychopathology, and psychotherapy}, vol.~5 (Academic
  press)
\bibAnnoteFile{plutchik2013emotion}

\bibitem[{Pope(1941)}]{pope1941importance}
Pope, A.~U. (1941).
\newblock The importance of morale.
\newblock \emph{The Journal of Educational Sociology} 15, 195--205
\bibAnnoteFile{pope1941importance}

\bibitem[{Rehurek and Sojka(2010)}]{rehurek2010software}
Rehurek, R. and Sojka, P. (2010).
\newblock Software framework for topic modelling with large corpora.
\newblock In \emph{In Proceedings of the LREC 2010 workshop on new challenges
  for NLP frameworks} (Citeseer)
\bibAnnoteFile{rehurek2010software}

\bibitem[{Rinker(2020)}]{qdap}
Rinker, T.~W. (2020).
\newblock \emph{{qdap}: {Q}uantitative Discourse Analysis Package}.
\newblock Buffalo, New York.
\newblock 2.4.2
\bibAnnoteFile{qdap}

\bibitem[{Roberts et~al.(2019)Roberts, Stewart, and Tingley}]{stm}
Roberts, M.~E., Stewart, B.~M., and Tingley, D. (2019).
\newblock {stm}: An {R} package for structural topic models.
\newblock \emph{Journal of Statistical Software} 91, 1--40.
\newblock \doi{10.18637/jss.v091.i02}
\bibAnnoteFile{stm}

\bibitem[{Shaver et~al.(1987)Shaver, Schwartz, Kirson, and
  O'connor}]{shaver1987emotion}
Shaver, P., Schwartz, J., Kirson, D., and O'connor, C. (1987).
\newblock Emotion knowledge: further exploration of a prototype approach.
\newblock \emph{Journal of personality and social psychology} 52, 1061
\bibAnnoteFile{shaver1987emotion}

\bibitem[{Silge and Robinson(2016)}]{tidytext}
Silge, J. and Robinson, D. (2016).
\newblock tidytext: Text mining and analysis using tidy data principles in r.
\newblock \emph{JOSS} 1.
\newblock \doi{10.21105/joss.00037}
\bibAnnoteFile{tidytext}

\bibitem[{{The Apache Software Foundation}(2009)}]{apache}
{The Apache Software Foundation} (2009).
\newblock \emph{Apache opennlp developer documentation}
\bibAnnoteFile{apache}

\bibitem[{Thelwall et~al.(2010)Thelwall, Wilkinson, and
  Uppal}]{thelwall2010data}
Thelwall, M., Wilkinson, D., and Uppal, S. (2010).
\newblock Data mining emotion in social network communication: Gender
  differences in myspace.
\newblock \emph{Journal of the American Society for Information Science and
  Technology} 61, 190--199
\bibAnnoteFile{thelwall2010data}

\bibitem[{Thet et~al.(2010)Thet, Na, and Khoo}]{thet2010aspect}
Thet, T.~T., Na, J.-C., and Khoo, C.~S. (2010).
\newblock Aspect-based sentiment analysis of movie reviews on discussion
  boards.
\newblock \emph{Journal of information science} 36, 823--848
\bibAnnoteFile{thet2010aspect}

\bibitem[{Tripto and Ali(2018)}]{tripto2018detecting}
Tripto, N.~I. and Ali, M.~E. (2018).
\newblock Detecting multilabel sentiment and emotions from bangla youtube
  comments.
\newblock In \emph{2018 International Conference on Bangla Speech and Language
  Processing (ICBSLP)} (IEEE), 1--6
\bibAnnoteFile{tripto2018detecting}

\bibitem[{Wickham et~al.(2022)Wickham, François, Henry, and Müller}]{dplyr}
Wickham, H., François, R., Henry, L., and Müller, K. (2022).
\newblock \emph{dplyr: A Grammar of Data Manipulation}.
\newblock Https://dplyr.tidyverse.org, https://github.com/tidyverse/dplyr
\bibAnnoteFile{dplyr}

\bibitem[{Yadollahi et~al.(2017)Yadollahi, Shahraki, and
  Zaiane}]{yadollahi2017current}
Yadollahi, A., Shahraki, A.~G., and Zaiane, O.~R. (2017).
\newblock Current state of text sentiment analysis from opinion to emotion
  mining.
\newblock \emph{ACM Computing Surveys (CSUR)} 50, 1--33
\bibAnnoteFile{yadollahi2017current}

\bibitem[{Yu and Wang(2015)}]{yu2015world}
Yu, Y. and Wang, X. (2015).
\newblock World cup 2014 in the twitter world: A big data analysis of
  sentiments in us sports fans’ tweets.
\newblock \emph{Computers in Human Behavior} 48, 392--400
\bibAnnoteFile{yu2015world}

\bibitem[{Zucco et~al.(2017)Zucco, Calabrese, and
  Cannataro}]{zucco2017sentiment}
Zucco, C., Calabrese, B., and Cannataro, M. (2017).
\newblock Sentiment analysis and affective computing for depression monitoring.
\newblock In \emph{2017 IEEE international conference on bioinformatics and
  biomedicine (BIBM)} (IEEE), 1988--1995
\bibAnnoteFile{zucco2017sentiment}

\end{thebibliography}

\end{document}